\documentclass[runningheads]{llncs}
\usepackage{graphicx}

\usepackage{tikz}
\usepackage{comment}
\usepackage{amsmath,amssymb} 
\usepackage{color}

\usepackage{epsfig}
\usepackage{microtype}
\usepackage{booktabs}

\usepackage{amsmath,amsfonts,bm}

\def\eqref#1{equation~\ref{#1}}

\def\1{\bm{1}}

\newcommand{\test}{\mathcal{D_{\mathrm{test}}}}

\def\vw{{\bm{w}}}
\def\vx{{\bm{x}}}

\DeclareMathAlphabet{\mathsfit}{\encodingdefault}{\sfdefault}{m}{sl}
\SetMathAlphabet{\mathsfit}{bold}{\encodingdefault}{\sfdefault}{bx}{n}

\DeclareMathOperator*{\argmin}{arg\,min}

\usepackage{url}
\usepackage{cite}
\usepackage[pagebackref=true,breaklinks=true,letterpaper=true,colorlinks,bookmarks=false]{hyperref}
\usepackage{subcaption}
\usepackage[ruled,vlined]{algorithm2e}
\usepackage{verbatim}
\usepackage{wrapfig}

\newcommand{\dknn}{Deep \textit{k}-NN }

\begin{document}
\pagestyle{headings}
\mainmatter
\def\ECCVSubNumber{15}  

\title{\dknn Defense Against Clean-label Data Poisoning Attacks} 

\titlerunning{\dknn Defense Against Data Poisoning Attacks}
%
\author{Neehar Peri\thanks{The first three authors contributed equally to this work} \and
Neal Gupta$^\star$ \and
W. Ronny Huang$^\star$ \and
Liam Fowl \and
Chen Zhu \and
Soheil Feizi \and
Tom Goldstein \and
John P. Dickerson}
\index{Huang, W. Ronny}
\index{Dickerson, John P.}
\authorrunning{N. Peri et al.}
%
\institute{Center for Machine Learning, University of Maryland - College Park
\email{\{wronnyhuang\}@gmail.com}, \email{\{john\}@cs.umd.edu}}
\maketitle

\begin{abstract}
Targeted clean-label data poisoning is a type of adversarial attack on machine learning systems in which an adversary injects a few correctly-labeled, minimally-perturbed samples into the training data, causing a model to misclassify a particular test sample during inference. Although defenses have been proposed for general poisoning attacks, no reliable defense for clean-label attacks has been demonstrated, despite the attacks' effectiveness and realistic applications. In this work, we propose a simple, yet highly-effective \dknn defense against both feature collision and convex polytope clean-label attacks on the CIFAR-10 dataset. We demonstrate that our proposed strategy is able to detect over 99\% of poisoned examples in both attacks and remove them without compromising model performance. Additionally, through ablation studies, we discover simple guidelines for selecting the value of \textit{k} as well as for implementing the \dknn defense on real-world datasets with class imbalance. Our proposed defense shows that current clean-label poisoning attack strategies can be annulled, and serves as a strong yet simple-to-implement baseline defense to test future clean-label poisoning attacks. Our code is available on \href{https://github.com/neeharperi/DeepKNNDefense}{GitHub}.

\end{abstract}

\section{Introduction}\label{sec:intro}

Machine-learning-based systems are increasingly being deployed in settings with high societal impact, including hate speech detection on social networks \cite{Rizoiu19:Transfer}, autonomous driving \cite{Chen17:Multi}, biometric-based applications \cite{Sun14:Deep}, and malware detection \cite{Pascanu15:Malware}. In these real world applications, a system's robustness to not only noise, but also \emph{adversarial manipulation} is paramount. With an increasing number of machine learning systems trained on data sourced from public and semi-public places such as social networks, collaboratively-edited forums, and multimedia posting services, adversaries can strategically inject training data to manipulate or degrade system performance.

\emph{Data poisoning} attacks on neural networks occur at training time, wherein an adversary places specially-constructed \emph{poisoned examples} into the training data with the intention of manipulating the behavior of the system at test time.  Recent work on data poisoning has focused on either (i) an attacker generating a small fraction of training inputs to degrade overall model performance, or (ii) a defender aiming to detect or otherwise mitigate the impact of that attack. In this paper, we focus on \emph{clean-label} data poisoning \cite{shafahi2018poison}, where an attacker injects a small number of \emph{correctly labeled}, minimally perturbed samples into the training data.  In contrast with traditional data poisoning, these samples are crafted to cause a model to misclassify a particular \emph{target} test sample during inference. These attacks are plausible in a wide range of applications, as they do not require the attacker to have control over the labeling function. Many large scale data sets are automatically scraped from the internet without direct supervision, so an adversary need only share their poisoned data online.


\emph{Our contribution}: In this paper, we initiate the study of defending against \emph{clean-label} poisoning attacks on neural networks by considering feature collision \cite{shafahi2018poison} and convex polytope attacks \cite{zhu2019transferable} on the CIFAR-10 dataset. Although poison examples are not easily detected by human annotators, we exploit the property that adversarial examples have different feature distributions than their clean counterparts in higher layers of the network, and that those features often lie near the distribution of the target class.  This intuition lends itself to a defense based on $k$ nearest neighbors in feature space, in which the poison examples are detected and removed \textit{prior} to training.
Further, the parameter $k$ yields a natural lever for trading off between the number of undetected poisons and number of discarded clean images when filtering the training set.

Our contributions can be outlined as follows.
\begin{itemize}
    \item We propose a novel \dknn defense for clean-label poisoning attacks. We evaluate it against state-of-the-art clean-label data poisoning attacks, using a slate of architectures and show that our proposed strategy detects 99\% of the poison instances without degrading overall performance.
    \item We reimplement a set of general data poisoning defenses \cite{koh2018stronger}, including $L_2$-Norm Outliers, One-Class SVMs, Random Point Eviction, and Adversarial Training as baselines and show that our proposed \dknn defense is more robust at detection of poisons in the trained victim models. 
    \item From the insights of two ablation studies, we assemble guidelines for implementing \dknn in practice. First we provide instructions for picking an appropriate value for $k$. Second, we provide a protocol for using the \dknn defense when class imbalance exists in the training set.
\end{itemize}

\section{Overview of Clean-label Data Poisoning}
\label{sec:overview}

We briefly describe the how clean-label data poisoning works and the intuition behind a neighborhood conformity defense. Figure \ref{fig:schem} shows the feature space representation (i.e. the representations in the penultimate layer of the network) for a targeted poisoning attack that causes a chosen target airplane image (feature representation shown as the dark gray triangle) to be misclassified as a frog during inference. To accomplish this, poison frog images (feature representation shown as dark orange circles) are perturbed to surround the target airplane in feature space. After training on this poisoned data set, the model changes its decision boundary between the two classes in order to accommodate the poison frogs, enveloping them onto the side of the frogs. Inadvertently, the nearby target airplane is also placed on the the side of the frogs, leading to misclassification. Under the \textit{feature collision} attack ~\cite{shafahi2018poison}, the perturbations are optimized so as to minimize the poison images' distance to the target image in feature space,

\begin{equation*}
    \vx_p = \argmin_{\vx}\; |\phi(\vx)-\phi(\vx_t)|^2_2 + |\vx-\vx_b|^2_2,
\end{equation*}

\noindent where $\vx_p$, $\vx_b$, $\vx_t$ are the poison, base, and target images, respectively, and $\phi$ is a feature extractor that propagates input images to the penultimate layer of the network. Alternatively, under the \textit{convex polytope} attack ~\cite{zhu2019transferable}, poisoned data points are optimized to form a convex hull of poisons around the target via a more sophisticated loss function. In both cases nonetheless, models fine-tuned on the poisoned dataset will have their decision boundaries adversarially warped and classify the targeted airplane image as a frog at inference time. Though the optimization causes a noticeable change in the feature representations of the images, the poison frogs are perturbed under some small $\ell_2$ or $\ell_\infty$ constraint so that they still appear to be frogs to a human observer. 

\subsection{Intuition behind \dknn Defense}
\begin{figure}[b!]
    \centering
    \includegraphics[width=0.75\textwidth, height=4cm]{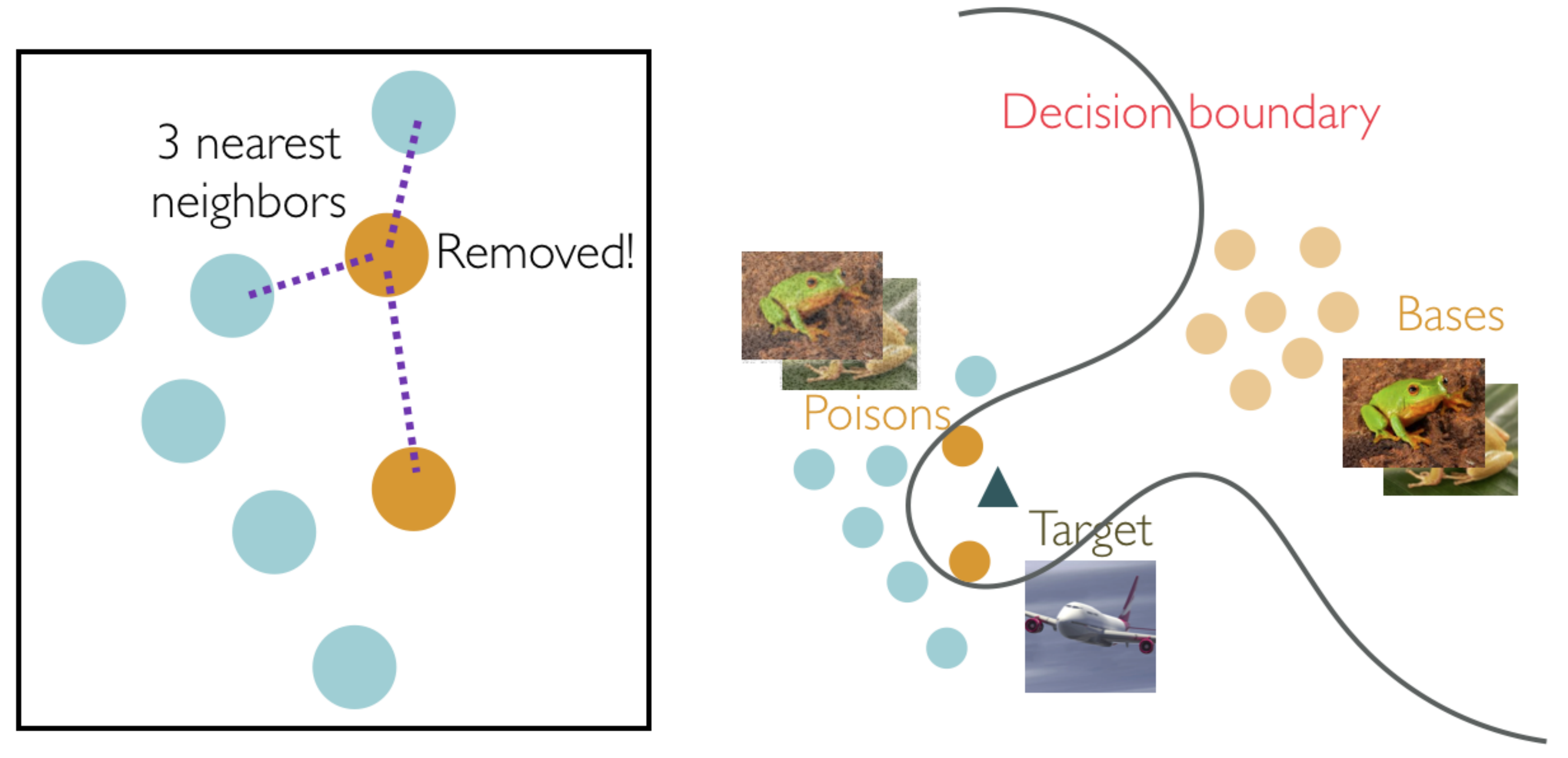}
    \caption{Proposed \dknn defense ($k=3$) correctly removing a poisoned example by comparing the class labels of poison with its $k$ neighbors. Since a majority of the $k$ points surrounding the poison do not share the same class label as the poison, it is removed.}
\label{fig:schem}
\end{figure}

As seen in Figure \ref{fig:schem}, poisons are surrounded by feature representations of the target class rather than of the base class. For instance, when $k=3$ and $n_{poison}=2$, each poison will almost always have a plurality of its neighbors as a non-poison in the target class. Since the plurality label of a poison’s neighbors does not match the label of the poison itself, the poison can be removed from the dataset or simply not used for training. More generally, if $k>2n_{poison}$, then we would expect the poisons to be outvoted by members of the target class and be filtered from the training set. Note that by setting $k > 2n_{poison}$, the poisons' label cannot be the majority, but may still be the plurality, or mode, of the \dknn set if the nearest neighbors of the current point are in multiple classes. Empirically, however, we do not observe this to be the case. Extracted features tend to be well-clustered by class; thus there are usually only 2 unique classes in the \dknn neighborhood, base class and target class, with the target class being larger. Therefore, in order to successfully defend against adversarial manipulation, a victim needs only to set a sufficiently large value of \textit{k} without needing to know exactly how many poisons there are \textit{a-priori}. We further elucidate on the effect of $k$ in Section \ref{sec:ablation}.
\section{Related Work}\label{sec:relwork}
We briefly overview related work in the space of defenses to adversarial attacks \cite{biggio2013evasion,FGSM}, which are categorized into two groups: inference time evasion attacks and train time data poisoning attacks.
Most adversarial defenses have focused on mitigating evasion attacks, where inference-time inputs are manipulated to cause misclassification. In neural networks, evasion adversarial examples are perturbed such that that the loss on the victim network increases. The search for an optimal perturbation is facilitated by use of the local gradient $\nabla_\textbf{x}\mathcal{L}$ obtained via backpropagation on either a white box network or a surrogate network if the victim network is unknown \cite{liu2016delving}. Many defenses against evasion attacks leverage the attacker's reliance on gradient information by finding ways to obfuscate gradients, using non-differentiable layers or reshaping the loss surface such that the gradients are highly uncorrelated. \cite{athalye2018obfuscated} showed that obfuscated gradient defenses are insufficient for defending against evasion attacks. Using various strategies to circumvent loss of gradient information, such as replacing non-differentiable layers with differentiable approximations during the backward pass, \cite{athalye2018obfuscated} demonstrates that stronger attacks can reduce inference accuracy to near zero on most gradient-based defenses. Defense strategies that withstand strong attacks are characterized by loss surfaces that are ``smooth'' with respect to a particular input everywhere in the data manifold. Variants of adversarial training \cite{madry2017towards, xie2019feature, shafahi2019adversarial} and linearity or curvature regularizers \cite{qin2019adversarial, moosavi2019robustness} have maintained modest accuracy despite strong multi-iteration PGD attacks \cite{madry2017towards}.

In evasion attacks, \dknn based methods have been used across multiple layers of a neural network to generate confidence estimates of network predictions as a way to detect adversarial examples \cite{papernot2018deep}. Similarly, \cite{sitawarin2019knn} proposes a white box threat model where an adversary has full access to the training set, and uses prior knowledge of model hyper-parameters, including the value of $k$ used in the \dknn defense when constructing poisons for general attacks. Our \dknn based defense differs in that it identifies and filters poisoned data at training time rather than at test time, using only ground truth labels. Furthermore, a soft nearest neighbor regularizer has been used during training time to improve robustness to evasion examples \cite{frosst2019analyzing}, but its resistance to clean-label poisoning examples has yet to be explored.

\textit{Backdoor} attacks have recently received attention from the research community as a realistic threat to machine learning models. Backdooring, proposed by \cite{gu2017badnets}, can be seen as a subset of data poisoning. In their simplest form, backdoor attacks modify a small number of training examples with a specific \textit{trigger} pattern that is accompanied by a \textit{target} label. These attacks exploit a neural network's ability to over fit to the training set data, and use the trigger at inference time to misclassify an example into the target class. The trigger need not change the ground truth label of the training example, making such attacks clean-label attacks \cite{turner2019cleanlabel}. However, these attacks rely upon the attacker being able to modify data at inference time, an assumption that may not always hold true, and one we do not make in this paper. A number of defenses to backdoor attacks have been proposed, primarily seeking to sanitize training data by detecting and removing poisons. Often, these defenses rely upon the heuristic that backdoor attacks create ``shortcuts'' in a neural network to induce target misclassification. \cite{steinhardt2017certified} employed two variants of an $\ell_2$ centroid defense, which we adapt in this paper. In one case, data is anomalous if it falls outside of an acceptable radius in feature space. Alternatively, data is first projected onto a line connecting class centroids in feature space and is removed based on its position on this line. 

\cite{chen2018detecting} proposed using feature clustering for data sanitation. This defense assumes that naive backdoor triggers will cause poison samples to cluster in feature space. The success for this defense diminishes drastically when exposed to stronger poisoning methods which do not use uniform triggers. Convex polytope attacks \cite{zhu2019transferable} create much stronger poisons by surrounding a target image in feature space with a convex hull of poisons. Such attacks will not always result in easily identifiable clusters of poisons. \cite{tran2018spectral} examines spectral signatures as a method for detecting backdoor attacks, stating that all attacks share a set of underlying properties. Spectral signatures are boosted in learned representations, and can be used to identify poisoned images through SVD.

\section{Defenses against Clean-Label Poisoning}
\label{sec:baseline-defenses}

In this section, we formally introduce the \dknn defense as well as a set of other baseline defenses against clean-label targeted poisoning attacks. We compare the effectiveness of each defense against both feature collision attacks and convex polytope attacks in Section \ref{sec:experiments}.

We use $\vx_t$ to denote the input space representation of the target image that an adversary tries to misclassify. The target has true label $l_t$ but the attacker seeks to misclassify it as having label $l_b$. We use $\vx_b$ to denote a base image having label $l_b$ that is used to build a poison after optimization. We use $\vx_w$ to denote a base image watermarked with a target image, that is $\gamma \cdot \vx_t +(1-\gamma)\cdot \vx_b$. To a human observer this image will retain the label $l_b$ when $\gamma$ is sufficiently low. We use $\phi(\vx)$ to denote the activations of the penultimate layer of a neural network. We refer to this as the \emph{feature layer} or \textit{feature space} and $\phi(\vx)$ as \emph{features} of $\vx$.

\noindent\textbf{Deep k-NN Defense}: For each data point in the training set, the \dknn defense takes the plurality vote amongst the labels of that point's $k$ nearest neighbors in feature space. If the point's own label is not the mode amongst labels of its $k$ nearest neighbors, the point is flagged as anomalous, and is not used when training the model. We use Euclidean distance to measure the distance between data points in feature space. See Algorithm \ref{alg:kNN}.
\begin{algorithm}[h]
\SetAlgoLined
\KwResult{Filtered training set $X^{train'}$}
 Let $S_k(x^{(i)})$ denote a set of $k$ points such that for all points $x^{(j)}$ inside the set and points $x^{(l)}$ outside the set,
 $|\phi(x^{(l)})-\phi(x^{(i)})|_2 \geq |\phi(x^{(j)})-\phi(x^{(i)})|_2$ \\
 $X^{\mathit{train}'}\leftarrow \{\}$ \\
 \For{Data points $\vx^{(i)} \in X^{\mathit{train}}$}{
  Let l denote the label of $x^{(i)}$ and let $l(S_k(x^{(i)}))$ denote the labels of the points in $S_k(x^{(i)})$ \\
  \eIf{$l \in mode(l(S_k(x^{(i)})))$}{
    $X^{\mathit{train}'}\leftarrow X^{\mathit{train}'} \cup \{x^{(i)}\}$;
   }{
   Omit $x^{(i)}$ from $X^{\mathit{train}'}$;
  }
 }
 \caption{\dknn Defense}
 \label{alg:kNN}
\end{algorithm}
 
\noindent\textbf{L2-Norm Outlier Defense}: The L2 norm outlier defense removes an $\epsilon > 0$ fraction of points that are farthest in feature space from the centroids of their classes. For each class of label $l\in\mathcal{L}$, with size $s_l = |{x^{(j)} \text{ s.t. } l(j) = l}|$, we compute the centroid $c_l$ as
\[
c_l = \frac{1}{s_l}\sum_{x^{(j)} s.t. l(j) = l} \phi(x^{(j)})
\]

\noindent and remove $\lfloor \epsilon s_l \rfloor$ points maximizing $|\phi(x^{(j)})-c_l|_2$. The L2 norm defense relies on the position of the centroid to filter outliers. However, the position of the centroid itself is prone to data poisoning if the per-class data size is small. This defense is adapted from traditional poison defenses not specific to neural networks \cite{koh2018stronger}.

%
%
%
%

\noindent\textbf{One-Class SVM Defense}: The one-class SVM defense examines the deep features of each class in isolation by applying the one-class SVM algorithm \cite{scholkopf2001estimating} to identify outliers in feature space for each label in the training set. It utilizes a radial basis kernel and is calibrated with a value $\nu=0.01$.

\noindent\textbf{Random Point Eviction Defense}: The random point eviction defense is a simple experimental control. It filters out a random subset of all training data. We remove 1\% of our training data for the feature collision attack and 10\% of our training data on the convex polytope attack. If the poisoning attack is sensitive to poisons being removed, the random defense may be successful, at the cost of losing a proportionate amount of the unpoisoned training data.

\noindent\textbf{Adversarial Training Defense}: Thus far, we have only considered defenses which filter out examples prior to training. We consider here another defense strategy that does not involve filtering, but rather involves an alternative victim training procedure. Adversarial training, used often to harden networks against evasion attacks \cite{goodfellow2015explaining, madry2017towards}, has been shown to produce neural network feature extractors which are less sensitive to weak features such as norm-bounded adversarial patterns  \cite{ilyas2019adversarial}. We explore here whether a victim's use of an adversarially trained feature extractor would yield features that are robust to clean-label data poisoning. Instead of the conventional loss over the training set, adversarial training aims to optimize

\begin{equation*}
    \underset{\theta}{\mathrm{min}}\; \mathcal{L}_\theta(X + \delta^*), \mathrm{where}\: \: \delta^*=\underset{\delta < \epsilon}{\mathrm{argmax}}\; \mathcal{L}_\theta(X + \delta),
    \label{eq:lower_level}
\end{equation*}

\noindent where $\theta$, $X$, and $\delta$ are the weights, training input, and adversarial perturbations, respectively, and $\mathcal{L}_\theta$ is some training loss (i.e., cross-entropy). In our experiments, we perform adversarial training following the standard procedure in \cite{madry2017towards}, using an $\ell_\infty$ PGD adversary of 20 steps and $\epsilon=8$.


\section{Evaluation}
\label{sec:experiments}

%

In this section, we evaluate the effectiveness of our \dknn defense and baseline defenses against the feature collision \cite{shafahi2018poison} and convex polytope \cite{zhu2019transferable} attacks on the CIFAR-10 dataset \cite{krizhevsky2009learning}. All model architectures, data splits, and hyperparameters are taken directly from the evaluation setups used in \cite{shafahi2018poison, zhu2019transferable}. 
We define the defense success rate as the number of times the poisoning attack fails to cause the target example to be misclassified, divided by the number of attempts. We only consider sets of poisons that lead to successful attacks in the undefended case so by definition the undefended defense success rate is 0\%.

%

\subsection{Defense against Feature Collision Attacks}

\subsubsection{Attack Procedure}
We randomly select 50 images in the base class. For each base image with input representation $\vx_b$, we compute the watermark base $\vx_w \leftarrow \gamma \cdot \vx_t +(1-\gamma)\cdot \vx_b$, then optimize $p$ with initial value $\vw$ using a forward-backward splitting procedure to solve
\[
\vx_p = \argmin_{\vx}|\phi(\vx)-\phi(\vx_t)|^2_2 + \beta |\vx-\vx_w|^2_2 
\] 
The hyperparameter $\beta$ is fixed at 0.1.
The resulting poisons $\vx_p$ are both close to the target image $\vx_t$ in feature space, and close to the watermarked input $\vx_w$ in image space. To ensure statistical significance, we craft 16 of these collections of 50 poisons and evaluate each collection independently.

\subsubsection{Defense Procedure}
As in the original setup \cite{shafahi2018poison}, we first train a modified AlexNet to convergence using only clean data. Next we apply our defenses on the set of clean data plus poisons to obtain a filtered dataset. That filtered dataset is then used to fine tune the pretrained model over 10 epoch with a batch size of 128. We evaluate the performance of all defenses described in Section~\ref{sec:baseline-defenses} against collections of 50 poisons that successfully cause a targeted misclassification. 

\subsubsection{Results}
As seen in Table~\ref{tab:transfer-poison-frogs}, the \dknn defense with $k=5000$, successfully identifies all but one poison across multiple attacks, while filtering just 0.6\% of the clean images from the training set. As a result, after victim training, models defended by \dknn have defense success rates of 100\%. In contrast, the $L2$-norm defense only identifies roughly half the feature collision poisons using $\epsilon=0.01$. Both the One-Class SVM and the Random Point Eviction defenses are unable to detect a majority of the feature collision poisons.

\begin{table*}[h!]
\centering
\caption{Comparing the effectiveness of baseline defenses aggregated for all model architectures in Feature Collision Attack}
\label{tab:transfer-poison-frogs}
\begin{tabular}{lp{1.5cm}p{2cm}p{2.2cm}p{2.2cm}}
\hline
Defense Strategy & Poisons Removed & Clean Images Removed (\%) & Defense Success Rate (\%) & CIFAR-10 Test Accuracy (\%) \\ \hline
\dknn ($k=5000$)& \textbf{799/800} & \textbf{0.6} & \textbf{100.0} & \textbf{74.6} \\
$L2$-Norm Outliers & 395/800 & 1.0 & 50.0 & \textbf{74.6} \\
One-class SVM & 168/800 & 1.0 & 37.5 & 74.5 \\
Random Point Eviction & 84/800 & 10.0 & 12.5 & 74.5           \\
\hline
\end{tabular}
\end{table*}


\subsection{Defense against Convex Polytope Attacks}
\subsubsection{Attack Procedure}

\label{sec:convex}
Following the procedure in \cite{zhu2019transferable}, the CIFAR-10 dataset is split into 48000 images for pretraining, and 500 images for fine-tuning. The poison base images are taken from the remaining split of 1500 images.

Since the attacker does not know the victim model parameters, they first pretrain their own model to convergence using the same subset of 48000 CIFAR-10 images used for pretraining. Next, an adversary uses this surrogate model to craft 5 poisons using the convex polytope method. To ensure statistical significance, 102 collections of 5 poisons are crafted.

When crafting convex polytope poisons, multiple surrogate models with different architectures are ensembled, so that the generated poisons generalize to victim architectures that the poisons were not crafted on. Our results are based on eight architectures: two of which are not used in crafting the poisons (black box setting), and six which use random initialization (grey box setting). The grey-box architectures are DPN92 \cite{chen2017dual}, GoogLeNet \cite{szegedy2015going}, MobileNetV2 \cite{sandler2018mobilenetv2}, ResNet50 \cite{he2016deep}, ResNeXT29-2x64d \cite{xie2017aggregated}, and SENet18 \cite{hu2018squeeze}, while the black-box architectures are DenseNet121 \cite{huang2017densely} and ResNet18 \cite{he2016deep}.

\subsubsection{Defense Procedure}
\begin{wrapfigure}{r}{0.5\linewidth}
\centering
\includegraphics[width=0.5\textwidth,  trim=0cm 1.5cm 0cm .2cm, clip, height=2.5cm]{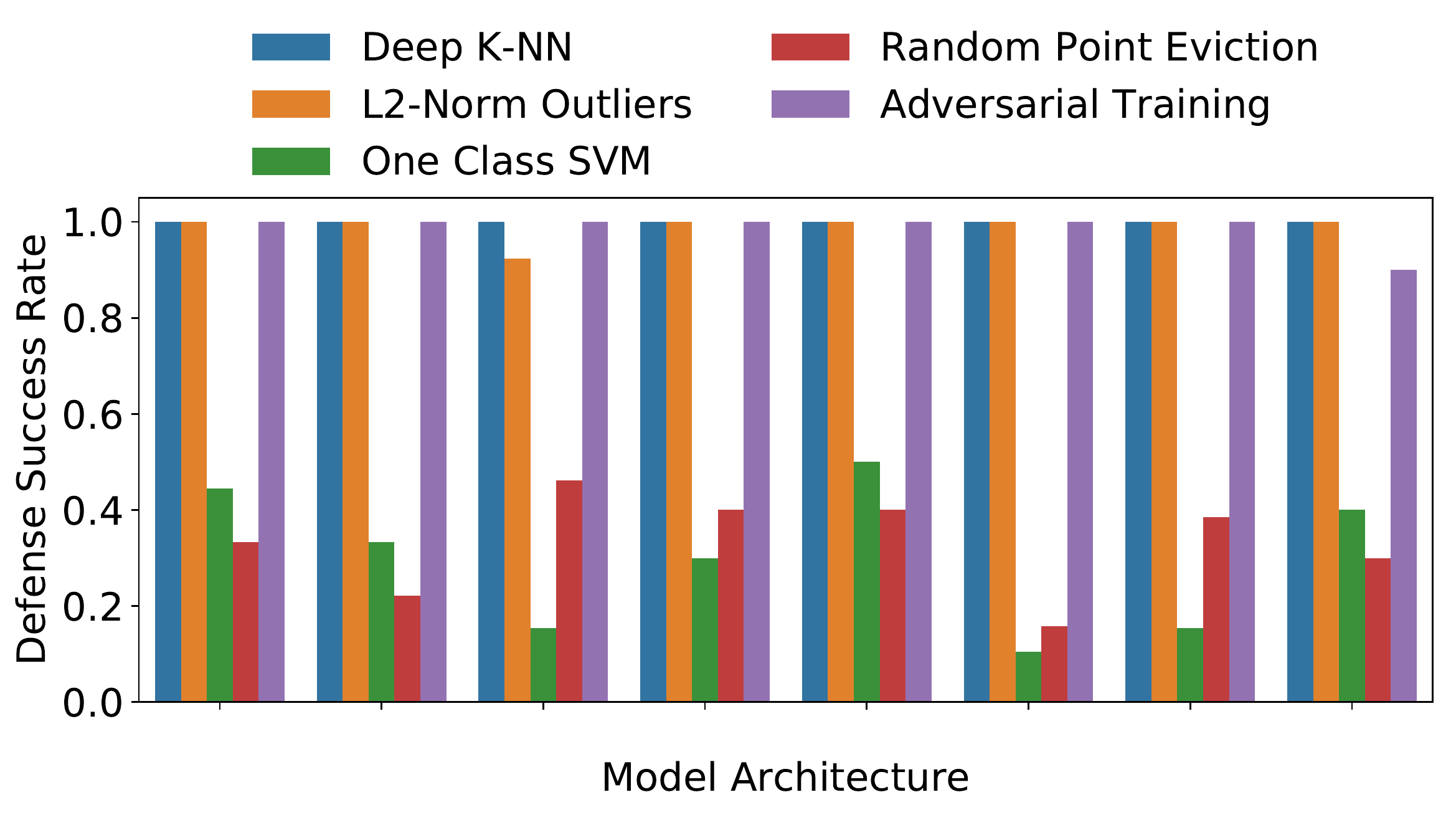} \\
\includegraphics[width=0.5\textwidth, height=3cm]{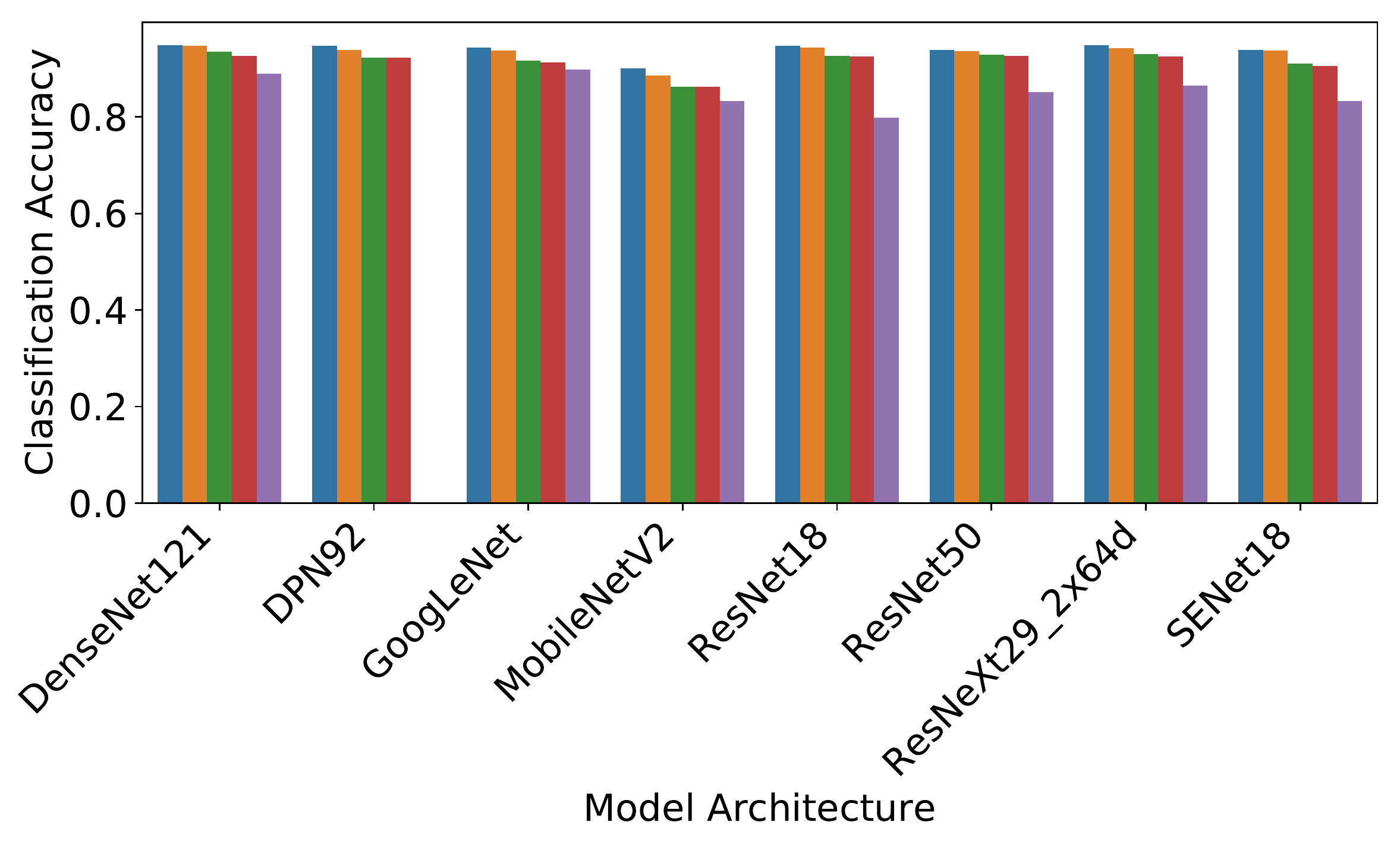}
\caption{The \dknn Defense is model-agnostic, achieving high defense success rate and test classification accuracy.}
\label{fig:arch}
\end{wrapfigure}

The victim model is first pretrained to convergence using a random initialization unknown to the attacker on the 48000 pretraining images from CIFAR-10\footnote{We use conventional training loss for all except the adversarial training defense.}. Our defenses are applied to the 500 fine-tuning images plus poisons to obtain a filtered fine-tuning set\footnote{There is no filtering in adversarial training.}. Finally, this filtered dataset is used to fine-tune the victim model.

Again, the performance of all defenses is reported only on collections of poisons that lead to a successful attack in the undefended case. Since the attacker did not have access to the victim architecture or model parameters during crafting of the poisons, the defenses are evaluated independently for each individual victim architecture.


\subsubsection{Results}
The aggregate results of each defense strategy on all 8 architectures are shown in Table~\ref{tab:transfer-convex}. Both the \dknn and $L2$-Norm defense filter out nearly all poisons, while incorrectly removing 4.3\% and 9.1\% of the clean training examples, respectively. Compared to feature collision poisons, convex polytope poisons trigger more false positive detections (i.e. clean images removed) across all defense methods, leading to fewer remaining clean examples and reduced test accuracy. 

Surprisingly, the $L2$-Norm defense is much better able to detect convex polytope poisons compared to feature collision poisons; it detects almost as many as \dknn. However, it has a lower specificity because it removes more clean images, resulting in half-percent lower test accuracy. These results are broken down for each victim architecture in Figure \ref{fig:arch}. The \dknn attack is successful on all architectures with perfect defense success rate. $L2$-norm Outliers and Adversarial Training perform almost as well. Other strategies largely fail to be a viable defense. 

We evaluate the effectiveness of adversarial training on the Convex Polytope-crafted poisons. In Table \ref{tab:transfer-convex} and Figure \ref{fig:arch}, adversarially trained feature extractors---trained naively to provide resistance against only evasion attacks---do in fact help mitigate poisoning attacks as well. To our knowledge, this is the first time adversarial training has been shown to provide resistance against data poisoning (i.e. training time) attacks and is a direction for future work.

\begin{wrapfigure}{r}{0.5\linewidth}
\centering
\includegraphics[width=0.42\textwidth, trim=.4cm 1.5cm 0cm .2cm, clip]{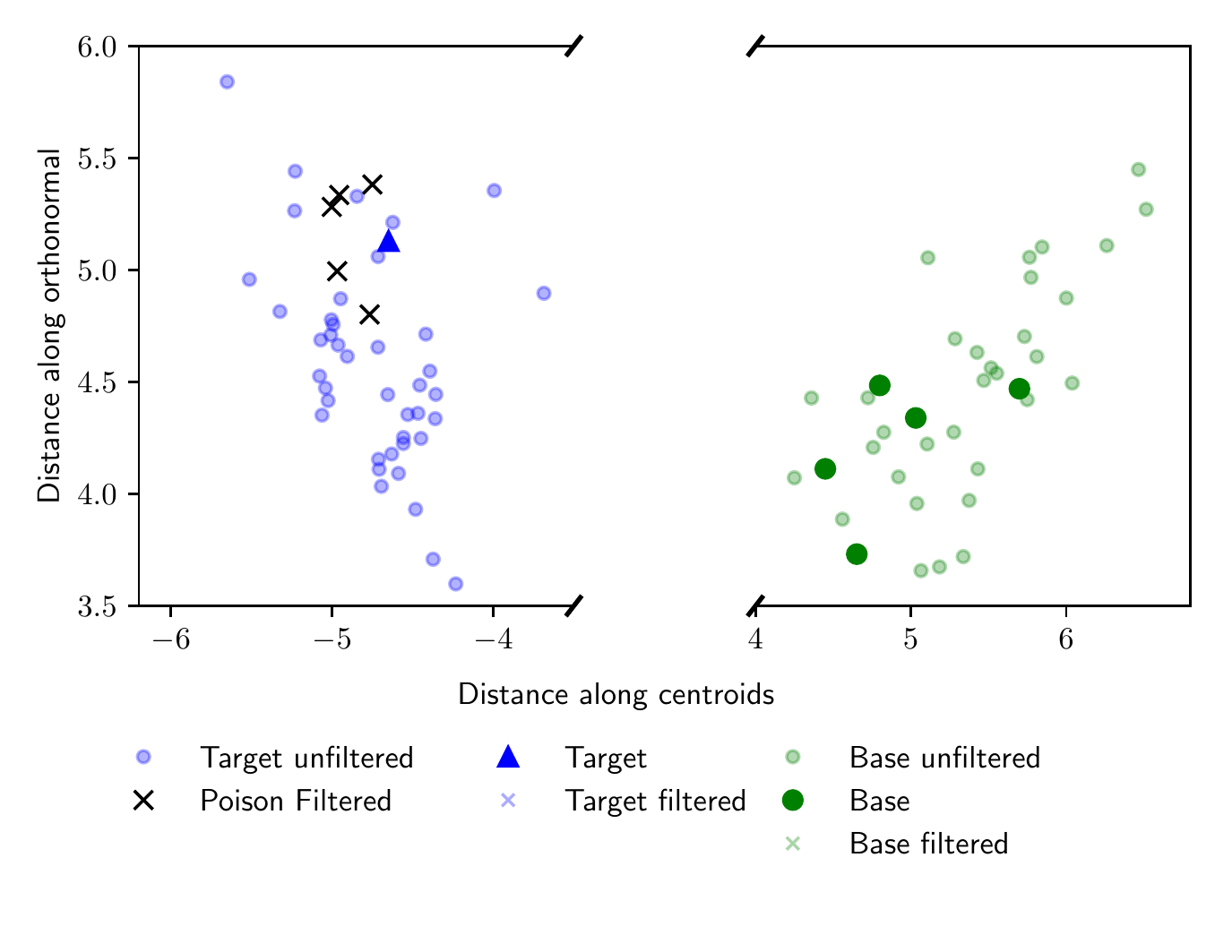}
\caption{Feature space visualization of the \dknn Defense against a Convex Polytope Attack on DPN92.}
\label{fig:transfer-poison-defense-visualization}
\end{wrapfigure}

The defense however significantly hurts test set accuracy (as is common for adversarially trained networks), which drops to 85\% on average, compared with 94\% on the same architectures without adversarial training. In scenarios when adversarial training for evasion attack robustness is not required, such as in situations when adversaries cannot control test time inputs, the \dknn defense provides the poisoning resistance without the burden of decreased generalization performance.

\begin{table*}[h]
\centering
\caption{Comparing the effectiveness of baseline defenses aggregated for all model architectures in Convex Polytope Attack}
\label{tab:transfer-convex}
\begin{tabular}{lp{1.5cm}p{2cm}p{2.2cm}p{2.2cm}}
\hline
Defense Strategy & Poisons Removed & Clean Images Removed (\%) & Defense Success Rate (\%) & CIFAR-10 Test Accuracy (\%) \\ \hline
\dknn $(k=50)$ & \textbf{510/510} & \textbf{4.3} & \textbf{100.0} & \textbf{93.9}\\
$L2$-Norm Outliers & 509/510 & 9.1 & 99.0 & 93.4 \\
One-class SVM  & 114/510 & 7.1 & 29.9 & 91.7 \\
Random Point Eviction & 47/510 & 10.0 & 33.2 & 91.3 \\
Adversarial Training & - & - & 98.6 & 85.2 \\
\hline
\end{tabular}
\end{table*}

\subsection{Feature Space Visualization}
The favorable results of \dknn defense also afford us an opportunity to understand anomaly detection in deep networks more generally via observing the effects in feature representations. A feature space visualization of the penultimate layer of the network is shown in Figure~\ref{fig:transfer-poison-defense-visualization}, with both filtered poisons and non-poisons displayed.

Specifically, Figure~\ref{fig:transfer-poison-defense-visualization} shows a projected visualization in the feature space of the fine tuning set in the target (blue) and base (green) classes.Following the projection scheme used in \cite{shafahi2018poison}, where the x-axis is the direction along the line connecting the centroids of the target and base class features and the y-axis is the component of the parameter vector (i.e. decision boundary) orthogonal to the between-centroids vector, the deep features of the DPN92 network are projected into a two-dimensional plane. The ``x" markers denote poisons that are filtered out by the defense and would have otherwise almost formed a convex polytope around the target (blue triangle). The \dknn acts with high specificity: all the poisons are filtered, while only 2 outlying clean points in the target class (not shown) are also filtered. No points in the base class are filtered.

\subsection{Limitations of the \dknn Defense}
The Deep k-NN defense exploits feature space clustering seen in feature collision and convex polytope attacks. It may not be as effective if this initial condition is not met. We view this as a strong and simple baseline defense for poisoning attacks that shows the need for more sophisticated and adaptive attacks.



\section{Ablation Studies and Best Practices}
\label{sec:ablation}

We now turn to ablation studies to gain insight into best practices for using the \dknn defense under realistic situations. All results are reported on the convex polytope attack for CIFAR-10 as described in \cite{zhu2019transferable} on all 8 architectures discussed previously. We specifically focus on the convex polytope attack method since it is shown to act as a stronger poison on black-box threat models, and study the transfer learning case to mimic the common practice of using pre-trained feature-extractors trained on large datasets.

\noindent We again closely mimick the setup in \cite{zhu2019transferable} using the first 4800 images in each class to train a model from scratch and then using the next 50 images of each class (making a fine-tuning set size of 500) to fine-tune the model. The Adam optimizer with a learning rate of 0.1 is used. In both studies, we assign frogs as the target class and ships as the base class. The first 5 ship images from the fine-tuning set are replaced with the 5 poisoned ships. Each set of 5 poisoned ships has an associated target frog image that is neither in the training nor fine-tune set. We use the standard CIFAR-10 test split to measure test accuracy. 

\subsection{Choosing a Value of \textit{k}}
In our first study, we vary the value of \textit{k} used in the \dknn defense. Since dataset sizes vary, as well as the number of classes, we normalize \textit{k} against the number of data points \textit{per class}. Specifically, we measure all metrics against a normalized-\textit{k} ratio, such that $\textrm{normalized-}k = k/N$ where \textit{k} is the number of nearest neighbors considered by the \dknn and \textit{N} is the maximum number of examples for any class in the fine-tune set.


\begin{figure*}[h!]
\centering
\includegraphics[width=0.48\textwidth]{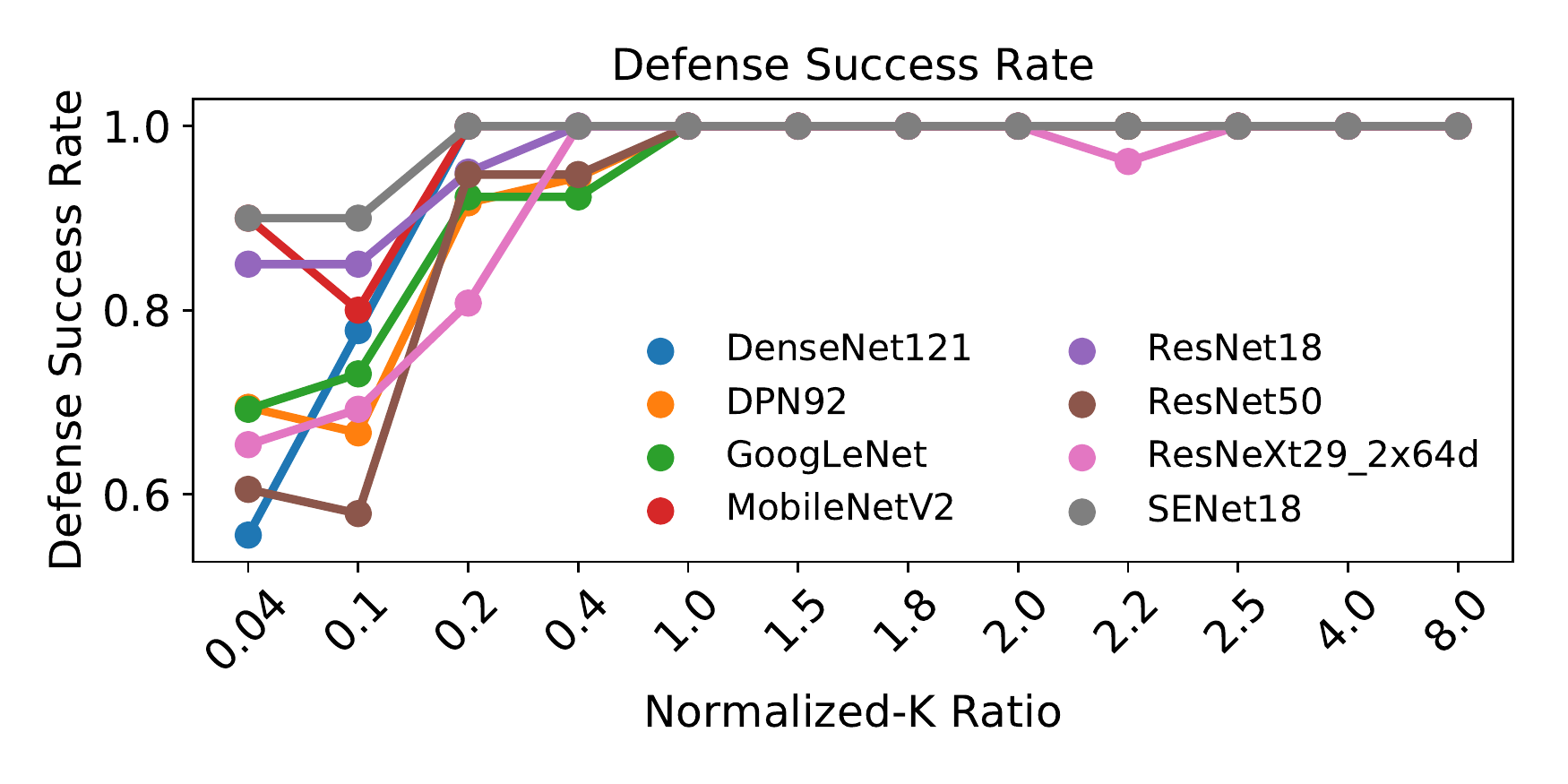}
\includegraphics[width=0.48\textwidth]{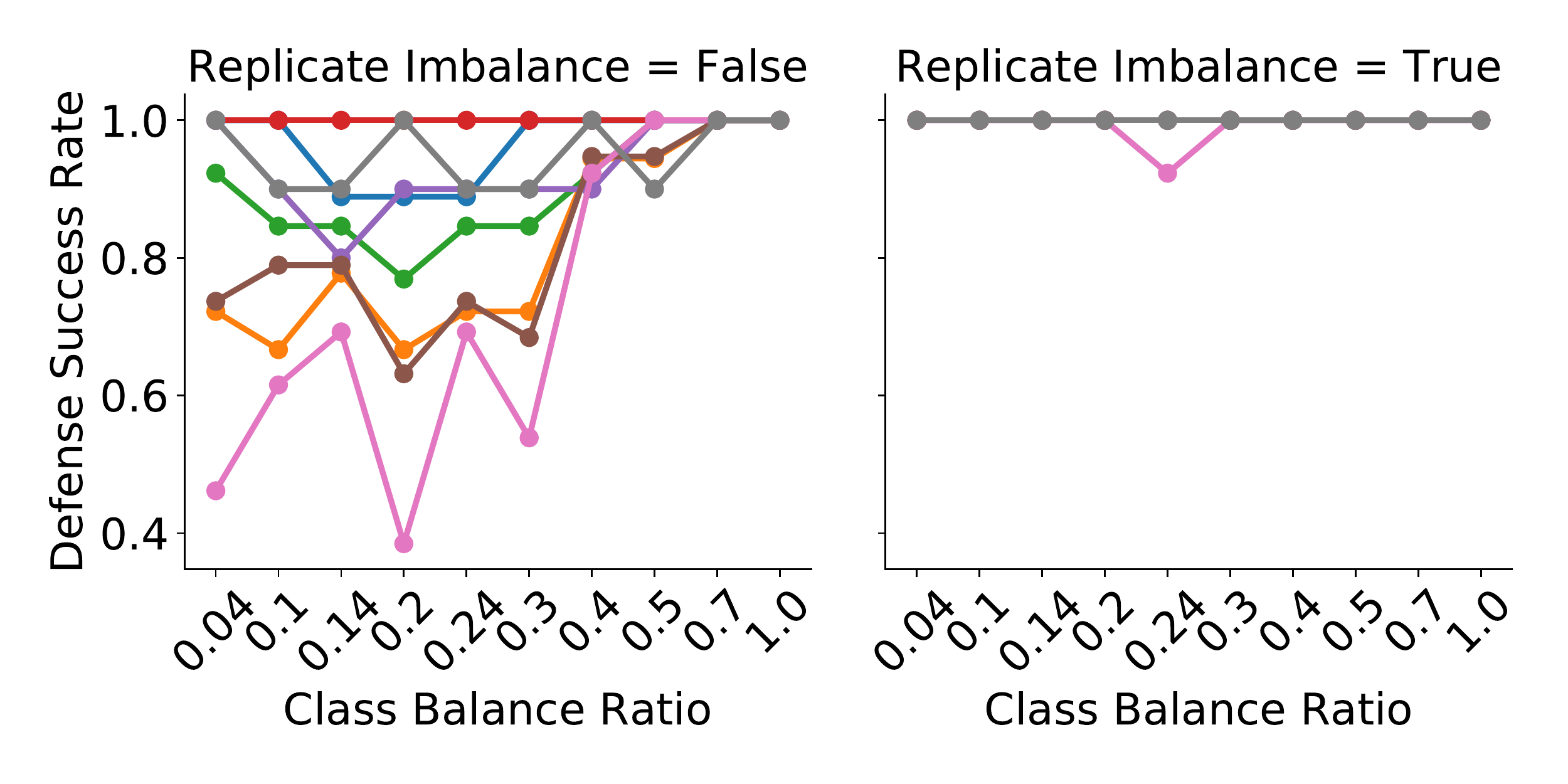} \\
\includegraphics[width=0.48\textwidth]{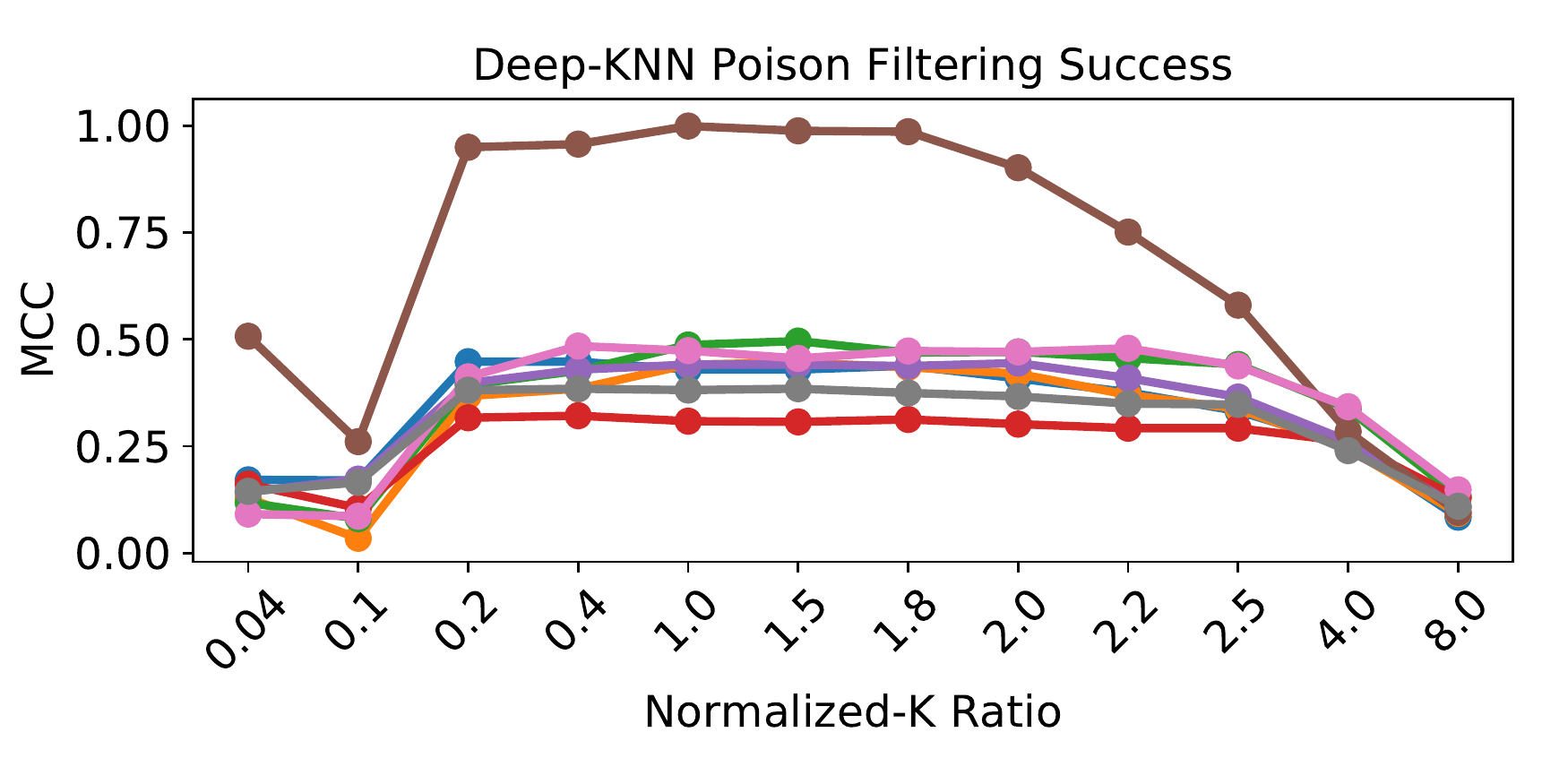}
\includegraphics[width=0.48\textwidth]{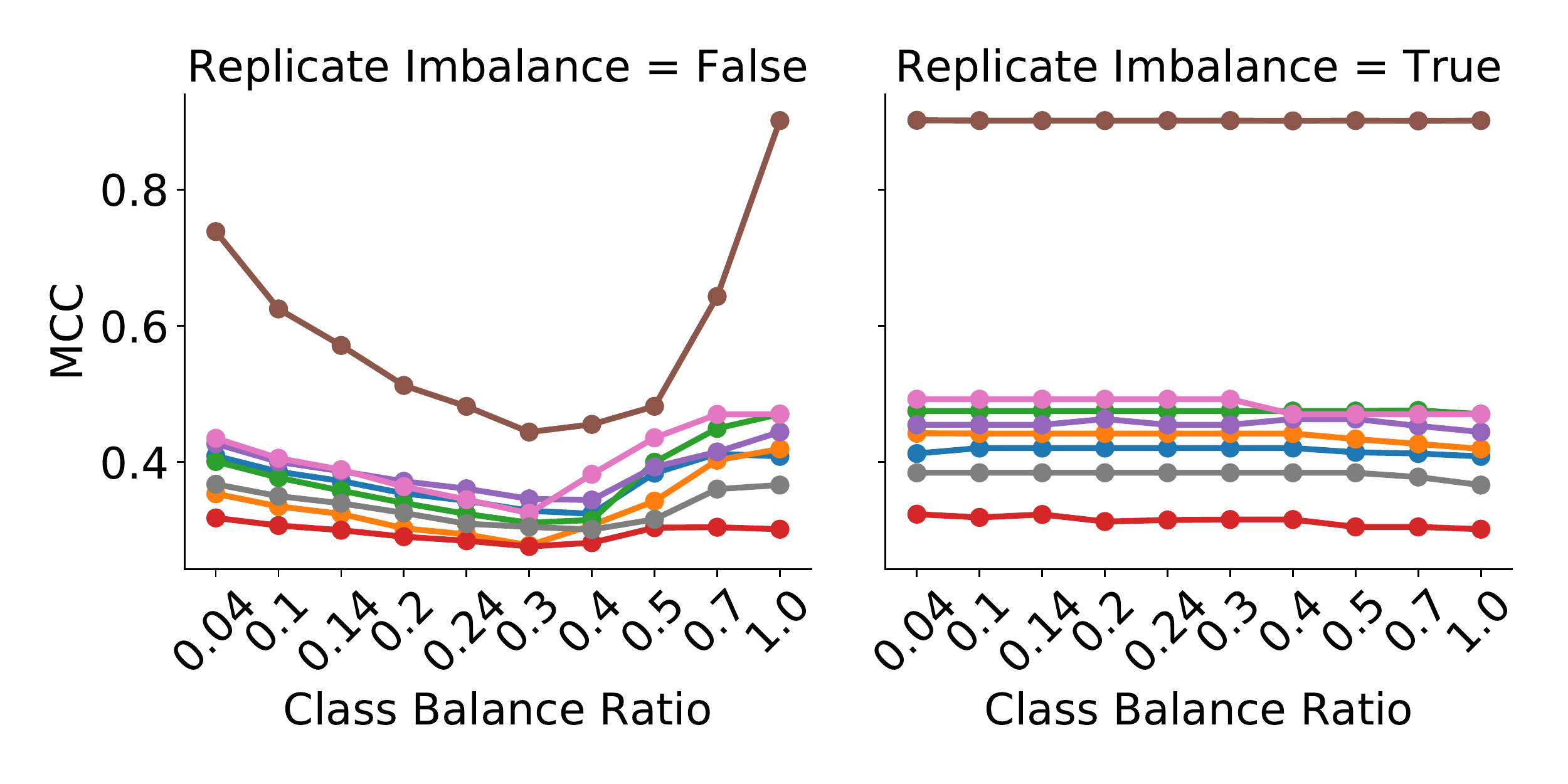} \\
\includegraphics[width=0.48\textwidth]{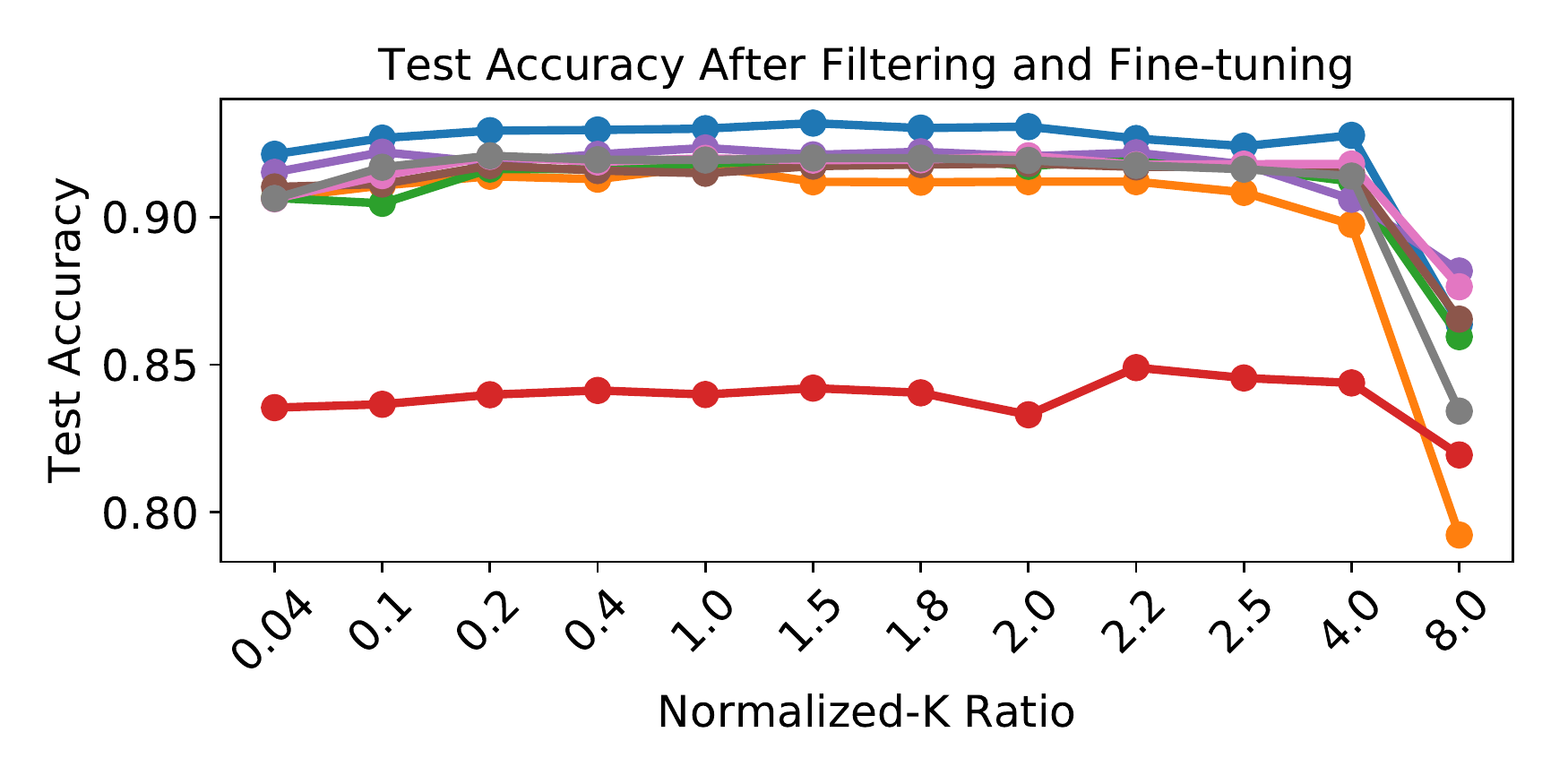}
\includegraphics[width=0.48\textwidth]{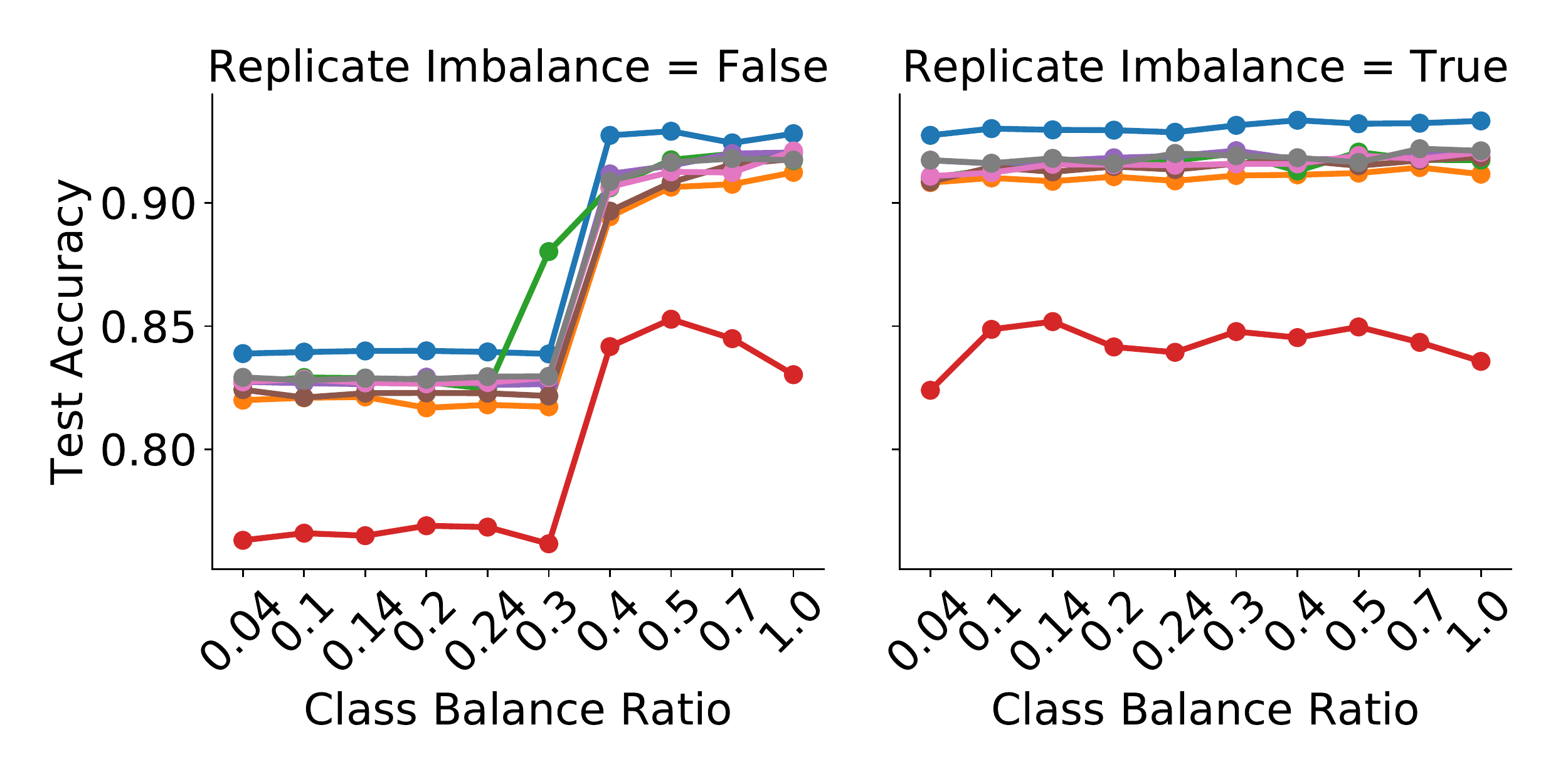} \\
\caption{Ablation studies on the effect of \textit{k} (left) and class imbalance (right). (Top Left) Defense success rate increases to 100\% for all models as normalized-\textit{k} ratio increases beyond 1.0 for all architectures. (Middle Left) Matthew's correlation coefficient is highest for all models when normalized-\textit{k} ratio is between 0.4 and 2.0. (Bottom Left) Accuracy on the CIFAR-10 test split drops as normalized-\textit{k} value increases beyond 4 times the number of examples per class. (Top Right) Defense and performance metrics under class imbalance. Defense success rate is stabilized when the target class training examples are first replicated to match the size of other classes. (Middle Right) Matthews correlation coefficient is also less dependent on the size of the target class when data replication is on. (Bottom Right) Test accuracy is highest when replicating the target examples to match the size of other classes.}
\label{fig:fixedClassBalance}
\end{figure*}

As seen in Figure \ref{fig:fixedClassBalance} (top left and middle left), the defense success rate and MCC begin to reach maximum levels at normalized-$k=0.2$, corresponding to an (unnormalized) $k$ of twice the number of poisons, $k=10=2n_{poison}$. This confirms our intuition in Section \ref{sec:overview}: when $k > 2n_{poison}$, poisons will be marked anomalous since the poison class cannot be the majority label in the neighborhood and is unlikely to be the plurality because the neighborhood usually only contains two unique classes. Of course, the victim must set a value of \textit{k} without knowledge of the number of poisons employed by the attacker. Fortunately we observe that defense success rate remains at 100\% as the normalized-\textit{k} ratio increases beyond $k=0.2$. Specifically, we see that after a normalized-\textit{k} value greater than 1.0 ($k=50$) (i.e. the situation where \dknn considers more neighbors than the per-class number of examples) the convex polytope attack is ineffective on all models. However, there are limitations. Despite successfully detecting all the poisons, an extremely large \textit{k} could lead to adverse effects on model test performance if too much clean data is removed (i.e. too many false positives). 

To take both positives and negatives into account, we again invoke the MCC metric in Figure \ref{fig:fixedClassBalance} (middle left) to measure the trade-off between detecting poisoned images and removing clean images. The maximum correlation coefficient for all models occurs for normalized-\textit{k} values in the range of 1 and 2. This makes intuitive sense. On one hand, for \textit{k} smaller than the class size, \dknn could fail to look within a large enough neighborhood around a data point to properly judge its conformity. For example, a poison point may lie within a small, yet very tight cluster of other poison points of the same class and be improperly marked as benign even though the poison cluster itself may lie within a much larger cluster of clean target points. On the other hand, for \textit{k} larger than 2 times the class size, the neighborhood may be too large and contain too many data points from a competing class. For example, the current target point may lie in a cluster of other target points, but since the neighborhood is so large that it contains all the target points as well as all the points in the nearby poison class cluster, the current target point will be improperly marked as anomalous. 


This upper threshold of normalized-$k=2$ is confirmed by looking at test accuracy performance in Figure \ref{fig:fixedClassBalance} (bottom left). We note that performance is highest in the normalized-$k$ region from 0.2 to 2. It slightly decreases after a normalized-\textit{k} ratio of 2 and sharply decreases after 4. This shows that a model's ability to generalize suffers when too many legitimate data points are removed under sufficiently large values of \textit{k}. Based on these experiments, we recommend using a normalized-\textit{k} value between 1 and 2 for optimal success in defending against poisoning attacks while minimizing false positives.

\subsection{Dealing with Class Imbalance}
In our second study, we consider the effectiveness of our defense on datasets with an imbalanced number of examples per class. Given an imbalanced dataset, the target class could be either the majority class or a minority class. The easiest case for the defender is when the target is the majority class. In this case, so long as \textit{k} is set sufficiently large, there will be more than enough target training examples to cause the poisons in their midst (in feature space) to be marked as anomalous after running \dknn. In this section, we will consider the worst case, wherein the target class is the smallest minority class in the dataset. Without applying any protocol to balance out the classes, there may not be enough target class neighbors when running \dknn to know that the poisons clustered in their midst are anomalous. 

A typical way to deal with imbalanced classes is to upweight the loss from examples in the minority classes or, equivalently, sample examples from minority classes at a higher rate that is inversely proportional to the fraction of the dataset that their class occupies. We consider a simple and equivalent modification of the latter protocol: given an imbalanced-class dataset, the examples in each class are \textit{replicated} by a factor of $N/n$, where $n$ is the number of examples in that class and $N$ is the maximum number of examples in any class. After this operation, the dataset will be larger, but once again balanced. We study the effect of this data replication protocol on imbalanced classes. Specifically, we set the number of examples in the target class (frog) to $n<N$ while leaving the number of examples in all other classes as $N$. We then replicate the frog examples by a factor of $N/n$ such that its size match the size of the other classes. Finally, we plot the defense success rate against the class imbalance ratio $n/N$ in Figure \ref{fig:fixedClassBalance} (top right). The value of normalized-\textit{k} is fixed at 2 ($k=100$) for this experiment.


Figure \ref{fig:fixedClassBalance} (top right, left panel) shows the defense success rate when no protocol is applied prior to running \dknn: the success rate suffers for class balance ratios below 0.7. When our data replication protocol is applied before the \dknn defense, the defense success rate is near perfect regardless of the class balance ratio. These results show that our minority class replication protocol, combined with the \dknn defense, is very effective at removing poisons in an imbalanced class dataset. Our replication-based balancing protocol normalizes the number of examples considered by the \dknn defense in feature space. 

Next, we observe the MCC as a function of class imbalance in the absence of any protocol in Figure \ref{fig:fixedClassBalance} (middle right, left panel). When the ratio is small, then the only thing that can hurt MCC is the misdetection of the targets as being anomalous. On the other hand, when the ratio is large, there is no class imbalance. MCC performs worst when there is a modest underrepresentation of the target class. That is where both the targets and the poisons can cause false negatives and false positives. When the replication protocol is applied in Figure \ref{fig:fixedClassBalance} (middle right), the MCC experiences an improvement, although the relative improvement is small. Interestingly, we observe that data replication stabilizes the MCC against class imbalance; the MCC is essentially a flat curve in Figure \ref{fig:fixedClassBalance} (middle right).



All models experience better test accuracy on the CIFAR-10 test set when replicating target examples as shown in Figure \ref{fig:fixedClassBalance} (bottom right). Despite only having $n$ \textit{unique} points in feature space, replicating them boosts model performance to be similar to the control experiment with a class balance ratio of 1.0. At lower class balance values, replicating data in unbalanced classes improves test accuracy by 8\%. Based on these experiments, we recommend the protocol of replicating images of underrepresented classes to match the maximum number of examples in any particular class prior to running \dknn. Defense success rate and model generalizability are both improved and stabilized by this protocol.
\section{Conclusion}

In summary, we have demonstrated that the simple \dknn approach provides an effective defense against clean-label poisoning attacks with minimal degradation in model performance. With an appropriately selected value of $k$, the \dknn defense identifies virtually all poisons from two state-of-the-art clean-label data poisoning attacks, while only filtering a small percentage of clean images. The \dknn defense outperforms other data poisoning baselines and provides a strong benchmark on which to measure the efficacy of future defenses. 
\section*{Acknowledgement}
Dickerson and Gupta were supported in part by NSF CAREER Award IIS-1846237, DARPA GARD HR00112020007, DARPA SI3-CMD S4761, DoD WHS Award HQ003420F0035, and a Google Faculty Research Award.  Goldstein and his students were supported by the DARPA GARD and DARPA QED4RML programs. Additional support was provided by the National Science Foundation DMS division, and the JP Morgan Fellowship program.
\bibliographystyle{splncs04}
\bibliography{egbib}

\end{document}